\pdfoutput=1

\documentclass[letterpaper, 10 pt, conference]{ieeeconf}  %

\IEEEoverridecommandlockouts                              %

\usepackage{graphics} %
\usepackage{epsfig} %
\usepackage{mathptmx} %
\usepackage{times} %
\usepackage{amsmath} %
\usepackage{amssymb}  %
\usepackage{comment}
\usepackage{algorithm,algpseudocode}
\usepackage{listings,xcolor}
\usepackage{subcaption}
\usepackage{booktabs}
\usepackage{amssymb}
\usepackage{array}
\usepackage{booktabs}
\usepackage{graphicx}
\usepackage{amssymb}
\usepackage{url}
\usepackage{adjustbox}
\usepackage{mathrsfs}
\usepackage{makecell}
\usepackage{pgfplots}

\usepackage{eso-pic}

\usepackage[hidelinks]{hyperref}
\hypersetup{pdftitle={Bridging Task Planning and Execution for Language-Conditioned Bimanual Manipulation},
            pdfauthor={Emanuele Cuzzocrea, Giuseppina Iannotti, Luca Rossini, Arturo Laurenzi, Nikos Tsagarakis}}
\pgfplotsset{compat=1.16}

\newcommand{\cmark}{\ensuremath{\checkmark}}
\newcommand{\xmark}{\ensuremath{\times}}

\renewcommand{\rothead}[1]{%
    \rotatebox[origin=bl]{45}{\scriptsize\bfseries #1}%
}

\title{\LARGE \bf
Bridging Task Planning and Execution for Language-Conditioned Bimanual Manipulation
}

\author{Emanuele Cuzzocrea$^{*,1,2}$, Giuseppina Iannotti$^{*,1,2}$,
Luca Rossini$^{1}$, Arturo Laurenzi$^{1}$, and 
Nikos Tsagarakis$^{1}$%
\thanks{$^{*}$Equal contribution.}%
\thanks{$^{1}$The authors are with the Humanoids and Human Centered
Mechatronics (HHCM), Istituto Italiano di Tecnologia, Genoa, Italy
(e-mail: {\tt\small \{name.surname\}@iit.it})}%
\thanks{$^{2}$The authors are with the University of Genoa, DIBRIS,
Genoa, Italy.}%
\thanks{\hphantom{$^{2}$}This work was supported by the European Union's
Horizon Europe research and innovation programme under grant agreement
No.~101070596 (euROBIN).}%
}

\begin{document}

\maketitle

\AddToShipoutPictureFG*{%
  \AtPageLowerLeft{%
    \put(\LenToUnit{\dimexpr1in+\hoffset+\oddsidemargin\relax},22){%
      \parbox{\textwidth}{%
        \fontsize{8}{9.5}\selectfont\leftskip=0pt plus .5fil\rightskip=0pt plus -.5fil\parfillskip=0pt plus 1fil\relax
        \textcopyright{} 2026 IEEE. Personal use of this
        material is permitted. Permission from IEEE must be
        obtained for all other uses, in any current or future
        media, including reprinting/republishing this
        material for advertising or promotional purposes,
        creating new collective works, for resale or
        redistribution to servers or lists, or reuse of
        any copyrighted component of this work in
        other works.%
      }%
    }
  }
}

\thispagestyle{empty}
\pagestyle{empty}

\begin{abstract}
Robots are increasingly expected to execute natural-language manipulation tasks in real-world environments. This is particularly relevant in household assistance and logistics, where clutter requires reasoning about both task objectives and execution feasibility. This challenge becomes even more pronounced for bimanual humanoid platforms, where coordinating two manipulators introduces additional interaction and kinematic constraints. We present a zero-shot vision-language framework for execution-aware bimanual manipulation that jointly reasons about task objectives and physical feasibility. Planning is grounded in a structured scene representation that combines semantic object relations with execution constraints, enabling the selection of obstacle relocations and execution choices while preserving task semantics. Compatible actions are organised into parallel bimanual stages with explicit pre- and post-conditions. A hierarchical validator evaluates these conditions under perceptual uncertainty and uses stage-level execution history to support local recovery or targeted replanning. 
Experiments on the CENTAURO hybrid wheeled-legged humanoid robot across nine language-conditioned manipulation scenarios show higher task success than a non-execution-aware baseline.
In dedicated failure scenarios, hierarchical validation recovers from execution failures with fewer model queries than always replanning.
These results demonstrate the benefits of combining execution reasoning directly into task planning.
Videos and additional material are available
on the \href{https://advrhumanoids.github.io/bridging-planning-execution/}{Project Webpage}.

\end{abstract}

\section{INTRODUCTION}
\label{sec:Introduction}

Robots operating in human-centred environments must translate natural-language instructions into physically executable manipulation behaviours~\cite{brohan2023can,huang2022inner,liang2023code}. This problem naturally arises in household assistance and logistics, where robots may need to retrieve requested objects, reorganise cluttered shelves, or manipulate items in dynamic workspaces. In these settings, each manipulation can alter object accessibility and the feasibility of subsequent actions, requiring continuous reasoning about both task objectives and execution constraints. This is particularly relevant in cluttered scenes, where retrieving a target object may require removing obstructions, selecting suitable relocation regions, and preserving the feasibility of later interactions~\cite{danielczuk2019mechanical,bejjani2021occlusion,huang2021visual}. Such constraints are especially important for humanoid robots performing bimanual manipulation, since coordinating two manipulators introduces additional interaction and kinematic dependencies~\cite{smith2012dual,krebs2022bimanual}. Task planning must therefore go beyond semantic reasoning and explicitly account for object accessibility, mutual obstruction, and robot kinematic limitations.

\begin{figure}
    \centering
    \includegraphics[width=1.0\linewidth]{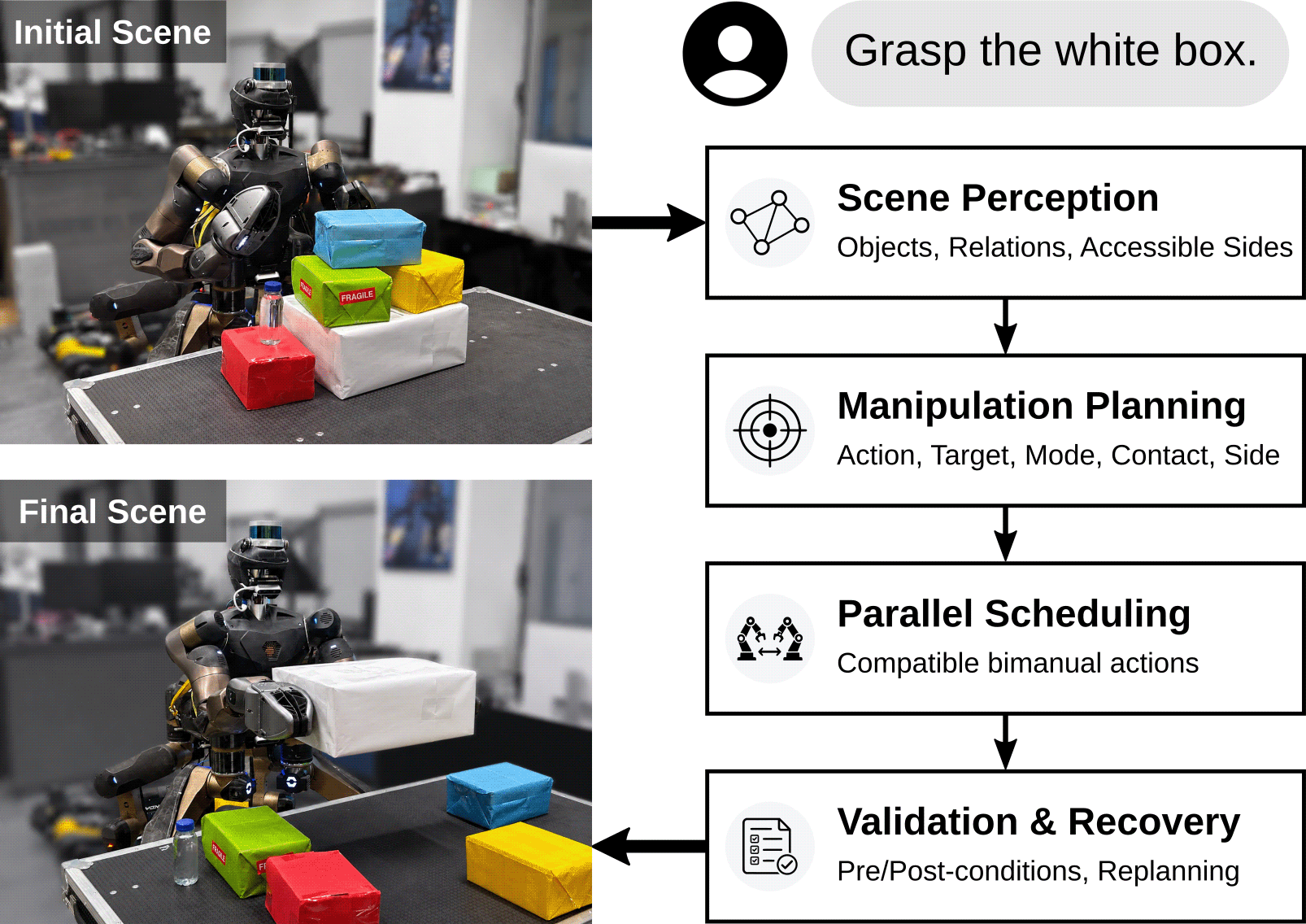}
    \caption{Execution-aware target retrieval in clutter. The framework builds
a structured scene representation, plans the manipulation actions and their
physical realisation, and schedules compatible actions for parallel bimanual
execution. A hierarchical validator monitors each execution stage and uses
structured feedback to support local recovery or targeted replanning.}
    \label{fig:teaser}
\end{figure}

Recent advances in foundation models have significantly improved language-conditioned robot planning. Vision-Language Models (VLMs) enable robots to interpret open-ended instructions and generate task-level manipulation plans~\cite{brohan2023can,singh2022progprompt,huang2022inner,liang2023code}, while Vision-Language-Action (VLA) policies directly predict robot actions from multimodal observations~\cite{zitkovich2023rt,team2024octo,kim2024openvla}. Despite these advances, existing approaches often either reason at a high semantic level or learn low-level behaviours end-to-end, providing limited support for explicitly reasoning about the physical interaction choices that determine whether a planned action can be realised.

Existing work addresses complementary aspects of this gap. Mechanical-search approaches explicitly reason about geometric obstruction and object rearrangement~\cite{danielczuk2019mechanical,bejjani2021occlusion,huang2021visual}, whereas planners based on structured scene representations capture semantic and spatial object relations~\cite{rana2023sayplan,gu2024conceptgraphs,lou2025explorevlm}. However, current formulations rarely represent the physical interaction choices required to bridge symbolic task plans and feasible manipulation. Consequently, accessibility constraints and interaction choices often remain implicit during planning, making it difficult to associate execution failures with the decisions that generated them.

Explicitly modelling physical interaction choices also provides a basis for structured execution validation. Existing systems can detect, explain, or recover from execution failures~\cite{guo2024doremi,liu2023reflect,duan2024aha}; however, when these choices remain implicit, feedback can reveal that an action has failed without identifying which planning requirement or expected effect was violated. Representing them through explicit conditions allows failures to be localised, execution progress to be assessed, and local recovery to be attempted before resorting to replanning.

We propose a zero-shot framework based on VLMs, i.e., requiring no task-specific training, fine-tuning, or demonstrations, that jointly reasons about task objectives and the physical interaction choices required to realise them in cluttered scenes, as illustrated in Fig.~\ref{fig:teaser}. This study builds upon our previous work~\cite{cuzzocrea2026executionaware}.
The main contributions are:
\begin{itemize}\itemsep0pt
\item a structured scene representation that jointly models semantic object relations, blocking dependencies, side-level accessibility, and robot reachability to support manipulation planning in cluttered environments;
\item an execution-aware planning framework that selects manipulation
actions, specifies their physical realisation, and schedules compatible
actions for parallel bimanual execution. A hierarchical validator
uses stage-level conditions and execution history to assess progress,
resolve perceptual uncertainty, and select local recovery or
targeted replanning;
\item an experimental evaluation on the CENTAURO hybrid wheeled-legged humanoid platform~\cite{kashiri2019centauro}, including real-world cluttered target-retrieval experiments, dedicated failure-recovery scenarios, and component ablation studies.
\end{itemize}

\begin{figure*}[t]
    \centering
    \includegraphics[width=1.0\linewidth]{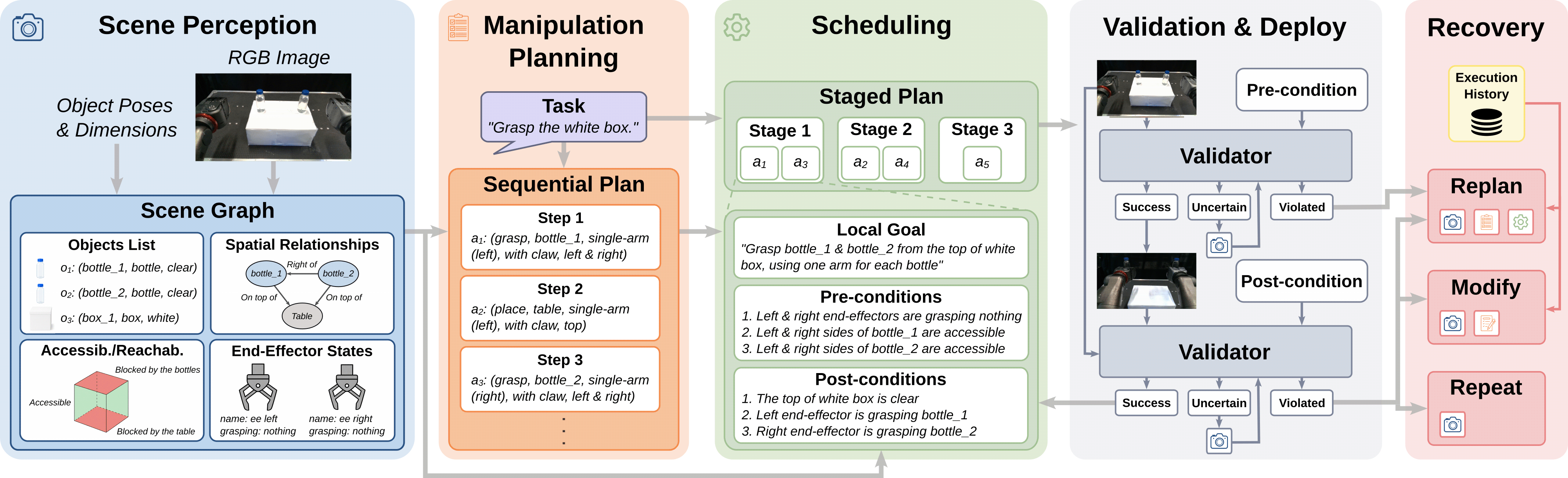}
    \caption{Overview of the proposed language-conditioned bimanual manipulation pipeline. From an RGB observation and object poses and dimensions, the framework constructs a structured scene representation, generates a sequential manipulation plan, and groups compatible actions into parallel execution stages. A hierarchical validator evaluates stage-level pre- and post-conditions, handles perceptual uncertainty, and leverages execution history to guide a local-to-global recovery process, prioritising local interventions whenever feasible.}
    \label{fig:scheme}
\end{figure*}

\section{Related Work}

Language-conditioned robotic manipulation spans several complementary research directions, including task planning, structured scene representation, bimanual coordination, and execution monitoring. Although recent methods increasingly ground high-level reasoning in robot observations and actions, the physical choices that connect task-level plans to feasible manipulation often remain implicit. This also limits the possibility of relating execution feedback to the planning decisions that generated it.

Early approaches to language-conditioned manipulation primarily focused on task-level planning. SayCan~\cite{brohan2023can} combines language reasoning with learned skill affordances to generate executable manipulation plans. Subsequent methods extend this paradigm through program generation and multimodal reasoning~\cite{singh2022progprompt,liang2023code,driess2023palm}. Related work on integrated task-and-motion planning explicitly connects symbolic decisions with motion feasibility~\cite{garrett2021integrated,siburian2025grounded}.
More recent methods further reduce the abstraction gap between semantic reasoning and robot execution. VoxPoser~\cite{huang2023voxposer} grounds manipulation objectives in 3D value maps, while ReKep~\cite{huang2024rekep} formulates manipulation through spatio-temporal relational constraints. Vision-Language-Action policies instead directly predict robot actions from multimodal observations~\cite{zitkovich2023rt,kim2024openvla,team2024octo,o2024open}. Despite this progress, the physical choices and conditions required to execute and validate individual manipulation actions often remain implicit.

One approach to reducing this gap is to ground planning in structured scene representations that organise the environment into semantic entities and spatial relations. Object-centric approaches~\cite{rana2023sayplan,gu2024conceptgraphs,lou2025explorevlm} capture semantic and spatial relationships relevant to task planning. Complementary work on mechanical search and object rearrangement~\cite{danielczuk2019mechanical,bejjani2021occlusion,huang2021visual} shows that successful manipulation also requires explicit reasoning about physical accessibility in cluttered environments, an aspect that remains challenging for current VLMs~\cite{fu2024blink,song2025robospatial}. These works highlight the importance of representing not only semantic relations but also the geometric constraints that determine how objects can be approached and manipulated.

Physical grounding becomes particularly important in bimanual manipulation, where actions must also be assigned and coordinated across two manipulators.
LABOR~\cite{chu2024large} uses language models to orchestrate sequential and concurrent skills, while LLM+MAP~\cite{chu2025llm+} combines language reasoning with symbolic multi-agent planning. However, arm assignment and concurrent execution are typically addressed through high-level skill orchestration or symbolic coordination, rather than being grounded in the same physical constraints used for manipulation planning.

\begin{table}[t]
    \centering
    \caption{Comparison of representative language-conditioned manipulation
    frameworks.}
    \label{tab:related_comparison}

    \begingroup
    \setlength{\tabcolsep}{1pt}
    \renewcommand{\arraystretch}{1.12}
    \footnotesize

    \begin{adjustbox}{max width=\columnwidth}
    \begin{tabular}{@{}l*{7}{c}@{}}
        \toprule

        \rule{0pt}{0.5cm}\textbf{Method}
        & \rothead{Zero-shot}
        & \rothead{\shortstack{Modular\\pipeline}}
        & \rothead{\shortstack{Explicit\\scene}}
        & \rothead{Bimanual}
        & \rothead{Accessibility}
        & \rothead{Reachability}
        & \rothead{Validation}
        \\

        \midrule

        SayCan~\cite{brohan2023can}
        & \cmark & \cmark & \xmark & \xmark
        & \xmark & \xmark & \xmark \\

        PaLM-E~\cite{driess2023palm}
        & \xmark & \xmark & \xmark & \xmark
        & \xmark & \xmark & \xmark \\

        LABOR~\cite{chu2024large}
        & \cmark & \xmark & \xmark & \cmark
        & \xmark & \cmark & \xmark \\

        LLM+MAP~\cite{chu2025llm+}
        & \cmark & \cmark & \cmark & \cmark
        & \xmark & \xmark & \xmark \\

        ExploreVLM~\cite{lou2025explorevlm}
        & \cmark & \cmark & \cmark & \xmark
        & \cmark & \xmark & \cmark \\

        \midrule

        \textbf{Ours}
        & \cmark & \cmark & \cmark & \cmark
        & \cmark & \cmark & \cmark \\

        \bottomrule
    \end{tabular}
    \end{adjustbox}

    \endgroup
\end{table}

Beyond planning and coordination, robust manipulation requires detecting deviations from the expected scene evolution and adapting the remaining strategy accordingly.
DoReMi~\cite{guo2024doremi}, REFLECT~\cite{liu2023reflect}, AHA~\cite{duan2024aha}, and ExploreVLM~\cite{lou2025explorevlm} support execution monitoring through misalignment detection, multimodal failure interpretation, or scene-based replanning.
However, these approaches generally do not link validation outcomes to the physical choices and stage-level conditions established during planning, limiting the ability to localise failures and select an appropriate recovery strategy.

Across these research directions, physical interaction choices
and execution verification conditions often remain implicit or
are represented separately. Table~\ref{tab:related_comparison}
compares representative approaches in terms of task-specific
training (\emph{Zero-shot}), modularity (\emph{Modular pipeline}),
structured object relations (\emph{Explicit scene}), dual-arm
coordination (\emph{Bimanual}), free approach directions
(\emph{Accessibility}), kinematic feasibility
(\emph{Reachability}), and online execution verification
(\emph{Validation}). Our framework explicitly integrates
interaction choices and verification conditions into bimanual
planning, combining physically grounded action specification,
parallel scheduling, and hierarchical stage-level validation.

\section{Method}
\label{sec:method}
Figure~\ref{fig:scheme} provides an overview of the proposed framework for language-conditioned bimanual manipulation in cluttered environments. Starting from a visual observation of the scene and a natural-language task instruction, the framework constructs an executable manipulation strategy while explicitly accounting for execution constraints and validating task progress throughout execution. The framework comprises four main components: structured scene representation, language-guided task planning, parallel execution scheduling, and hierarchical execution validation. The individual components are described in the following sections.

\subsection{Object Geometric Representation}
\label{subsec:domain}

Objects are represented as oriented bounding boxes defined by their position, dimensions, and orientation. Each bounding box comprises three pairs of opposing parallel faces (\emph{left-right}, \emph{front-back}, and \emph{top-bottom}), whose labels are assigned with respect to the robot reference frame and therefore describe workspace-relative approach directions. In particular, the front face is defined as the face oriented towards the robot, while the remaining labels follow accordingly. The proposed framework assumes objects that can be grasped through opposing contacts on one pair of parallel faces, allowing grasp accessibility and object interactions to be inferred directly from geometry. Objects requiring instance-specific grasp affordances, such as a mug that must be grasped by its handle, fall outside the scope of the framework. This is a deliberate scoping choice: many rigid objects commonly encountered in tabletop and logistics settings can be approximated through this geometric representation, and these are precisely the scenarios in which mutual obstruction constitutes the dominant difficulty addressed here.

\subsection{Structured Scene Representation}
\label{subsec:scene}

Given an RGB image $I_t$ acquired from the robot viewpoint at time $t$, scene construction is performed in two stages. First, a VLM processes $I_t$ to identify the object instances in the scene, infer their semantic attributes and qualitative pairwise relations, and determine the interaction state of the robot end-effectors. This produces a symbolic description of the scene. Second, FoundationPose~\cite{wen2024foundationpose}, in its model-based mode with CAD models of the objects, provides 6D object pose estimates that geometrically ground the detected objects and their relations. This geometric information is used to define the six robot-relative sides of each object and enrich the symbolic representation with the interaction constraints described below.

The scene is represented as a graph
\begin{equation}
    G_t = (V_t, R_t),
    \label{eq:graph}
\end{equation}
where $V_t = O_t \cup E_t$ is the set of scene entities, composed of object
nodes $O_t$ and end-effector nodes $E_t$, and $R_t$ is the set of pairwise
relations between them. Objects are represented as discrete instances with
semantic attributes such as category and colour, together with any property
relevant to manipulation, such as fragility. End-effectors are instead modelled
through their current interaction state, namely whether they are empty or
holding an object. Unlike objects, they are not associated with persistent
scene properties, since their pose depends on the current robot configuration.

The scene structure inferred by the VLM is captured through
pairwise relations
\begin{equation}
    R_t \subseteq V_t \times T \times V_t,
    \label{eq:relations}
\end{equation}
where $T$ defines the set of admissible relation types, namely
\emph{on top of}, \emph{right of}, \emph{in front of}, and
\emph{grasped by}. Entity identities and pairwise relations provide a structural description of
the scene, but do not by themselves specify how the objects can be physically
interacted with. The symbolic graph is therefore enriched using the estimated
object poses and dimensions. Specifically, the six sides of each object are
annotated with their interaction status in terms of \emph{accessibility} and
\emph{reachability}. \emph{Accessibility} expresses whether sufficient free space is available around a
side for the end-effector to interact with it. For each spatial relation inferred
in the scene graph, the distance separating the corresponding facing sides of
the two objects is computed from their estimated poses and dimensions. This
distance is compared with a safety threshold derived from the end-effector
geometry. When a side is blocked, the
object or support surface responsible for the obstruction is also recorded, so
that each unavailable interaction direction is represented together with its
cause. \emph{Reachability} is evaluated separately, since geometric accessibility does not
guarantee that a side can be approached given the robot kinematics. For each candidate interaction direction, reachability is determined by
checking whether the corresponding approach pose admits a feasible inverse
kinematics solution.

\subsection{Manipulation Planning}
\label{subsec:planning}

\begin{table}[t]
\centering
\caption{Modules re-invoked by each recovery strategy, ordered from local to
global.}
\label{tab:recovery}
\begingroup
\setlength{\tabcolsep}{5pt}
\renewcommand{\arraystretch}{1.12}
\footnotesize
\begin{tabular}{lcccc}
\toprule
\textbf{Strategy} & \textbf{Scene} & \textbf{Action} & \textbf{Manip.}
& \textbf{Scheduler} \\
& \textbf{Perception} & \textbf{Revision} & \textbf{Planning} & \\
\midrule
\emph{repeat}  & \cmark & \xmark & \xmark & \xmark \\
\emph{modify}  & \cmark & \cmark & \xmark & \xmark \\
\emph{replan}  & \cmark & \xmark & \cmark & \cmark \\
\bottomrule
\end{tabular}
\endgroup
\end{table}

Given the structured scene representation $G_t$, the natural-language task
instruction $H_t$, a VLM generates an ordered manipulation plan
\begin{equation}
    \Pi_t = (a_1, a_2, \dots, a_n),
    \label{eq:plan}
\end{equation}
which describes a sequence of scene transformations that progressively
establishes the conditions required to complete the task. Each action is
represented as
\begin{equation}
    a_k = (\textit{action}_k, \textit{target}_k, \textit{mode}_k,
           \textit{contact}_k, \textit{side}_k),
    \label{eq:action}
\end{equation}
where $\textit{action}_k$ denotes the manipulation primitive, chosen among
\emph{grasp}, \emph{place}, \emph{push}, and \emph{release};
$\textit{target}_k$ identifies the manipulated object or the placement
destination; $\textit{mode}_k$ specifies whether the action is performed with
the left arm, the right arm, or both arms jointly; $\textit{contact}_k$
indicates whether the interaction is performed with or without engaging the
claw; and $\textit{side}_k$ identifies the object side, or pair of opposing
sides, from which the interaction should be executed. This representation connects task-level decisions with their physical
realisation. For instance, a grasp must involve an accessible pair of opposing sides reachable by the selected arm, whereas a push specifies the side on which contact is established. When the task target is not directly graspable, the planner reasons about the recorded causes of obstruction. It first considers alternative accessible grasp directions and, when none is feasible, introduces actions to relocate the objects blocking the required sides. Each relocation is defined relative to a reference object and must place the obstacle far enough from it to satisfy the same end-effector clearance threshold used to determine accessibility. The use of a VLM also allows semantic constraints expressed in the instruction or encoded in the scene representation, such as object fragility, to influence both the selected action and the modality of interaction.

\begin{figure}[t]
    \centering
    \begin{subfigure}[t]{0.32\linewidth}
        \centering
        \includegraphics[width=\linewidth]{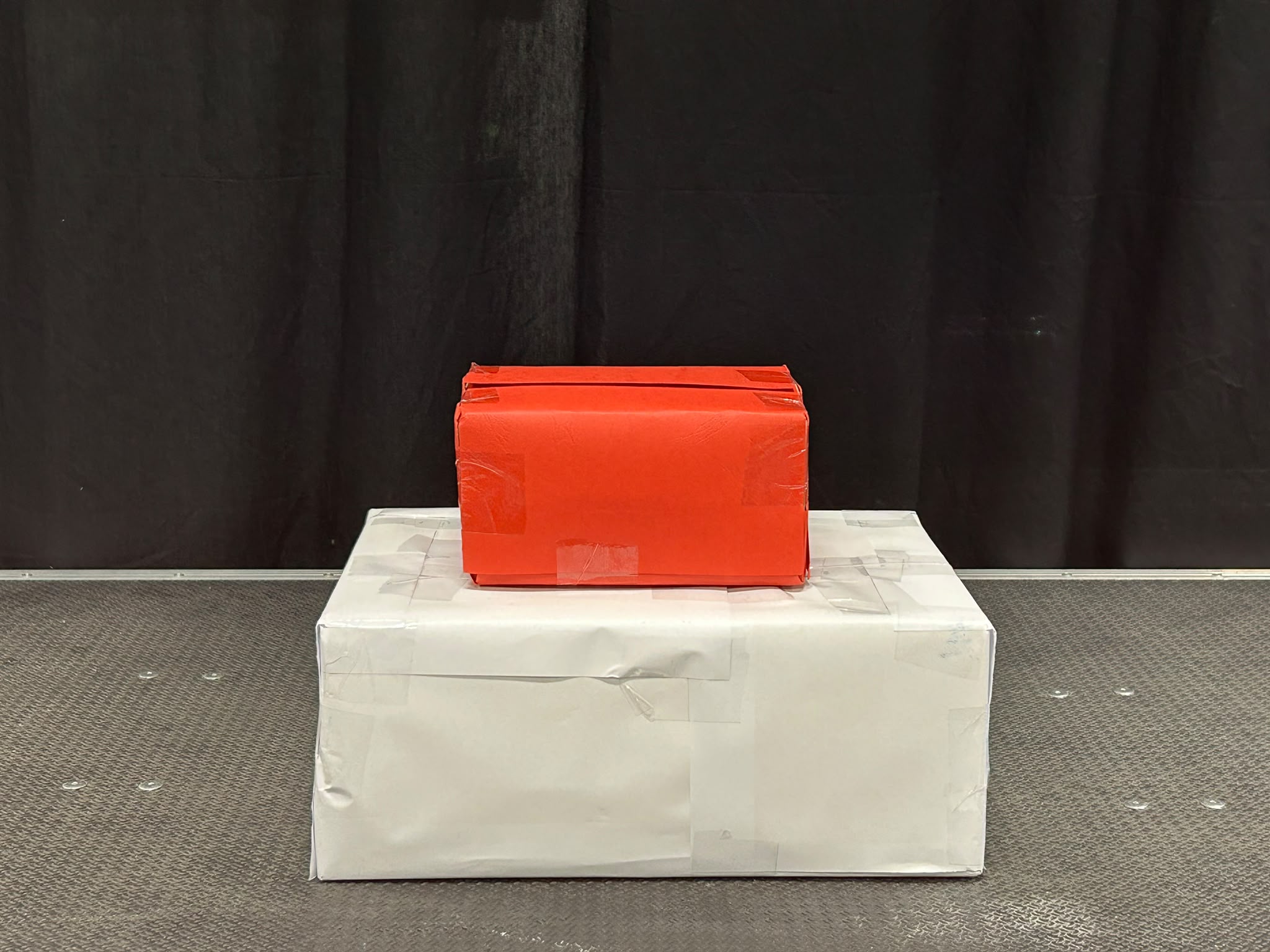}
        \caption{Stacked Boxes}
    \end{subfigure}
    \hfill
    \begin{subfigure}[t]{0.32\linewidth}
        \centering
        \includegraphics[width=\linewidth]{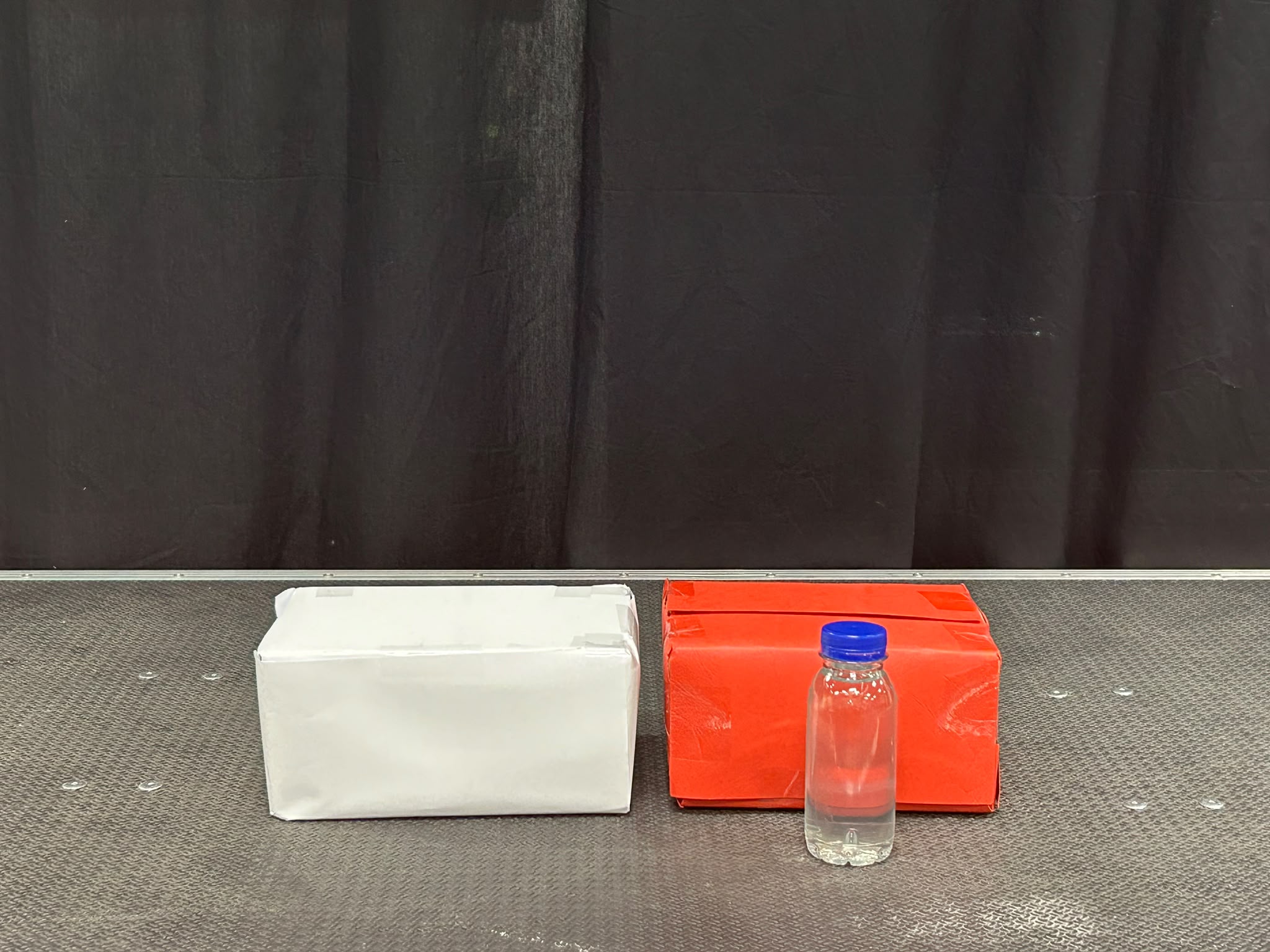}
        \caption{Boxes Aside}
    \end{subfigure}
    \hfill
    \begin{subfigure}[t]{0.32\linewidth}
        \centering
        \includegraphics[width=\linewidth]{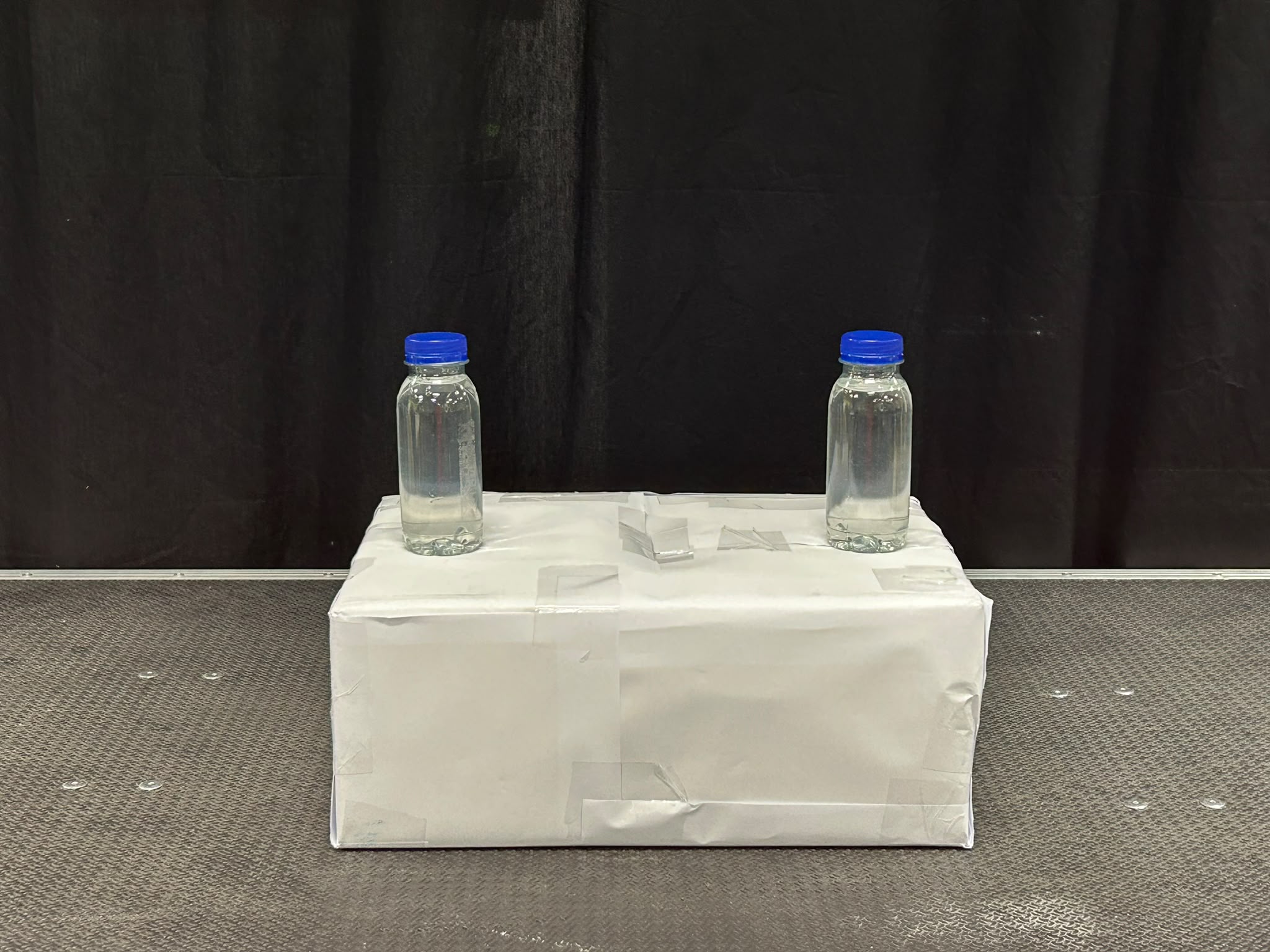}
        \caption{Bottles on Box}
    \end{subfigure}

    \vspace{0.15cm}

    \begin{subfigure}[t]{0.32\linewidth}
        \centering
        \includegraphics[width=\linewidth]{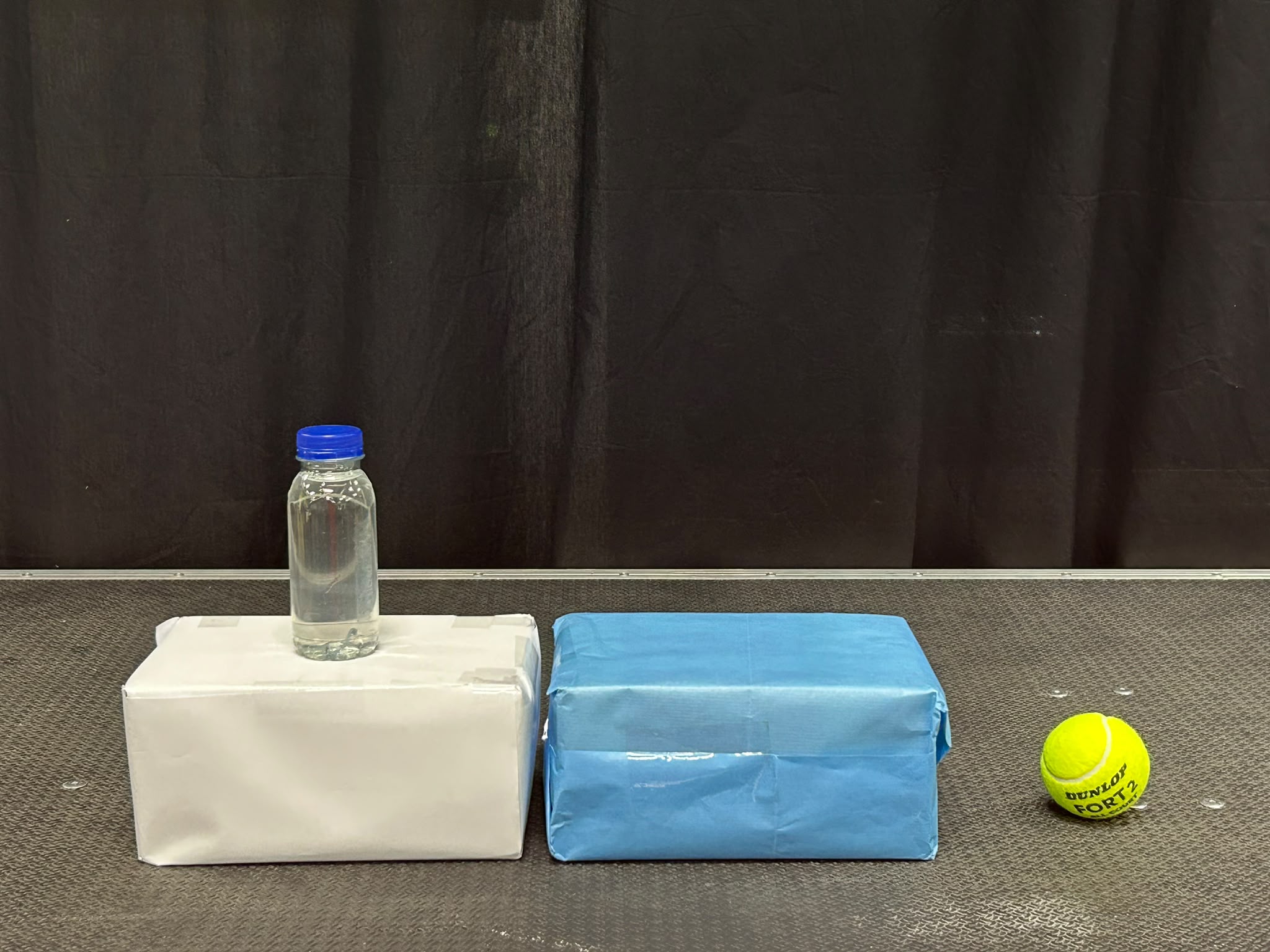}
        \caption{Scene Update}
    \end{subfigure}
    \hfill
    \begin{subfigure}[t]{0.32\linewidth}
        \centering
        \includegraphics[width=\linewidth]{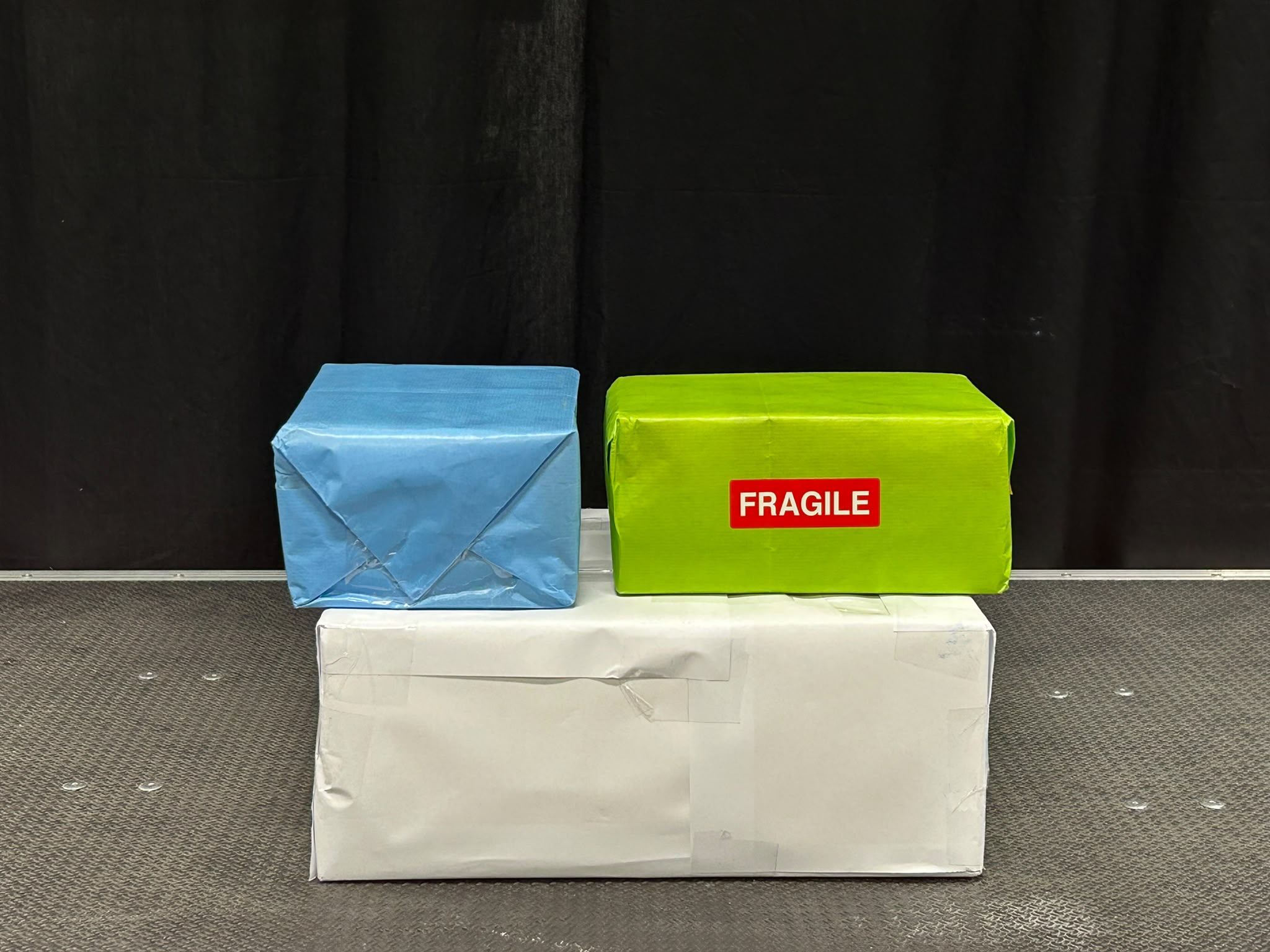}
        \caption{Fragile Box}
    \end{subfigure}
    \hfill
    \begin{subfigure}[t]{0.32\linewidth}
        \centering
        \includegraphics[width=\linewidth]{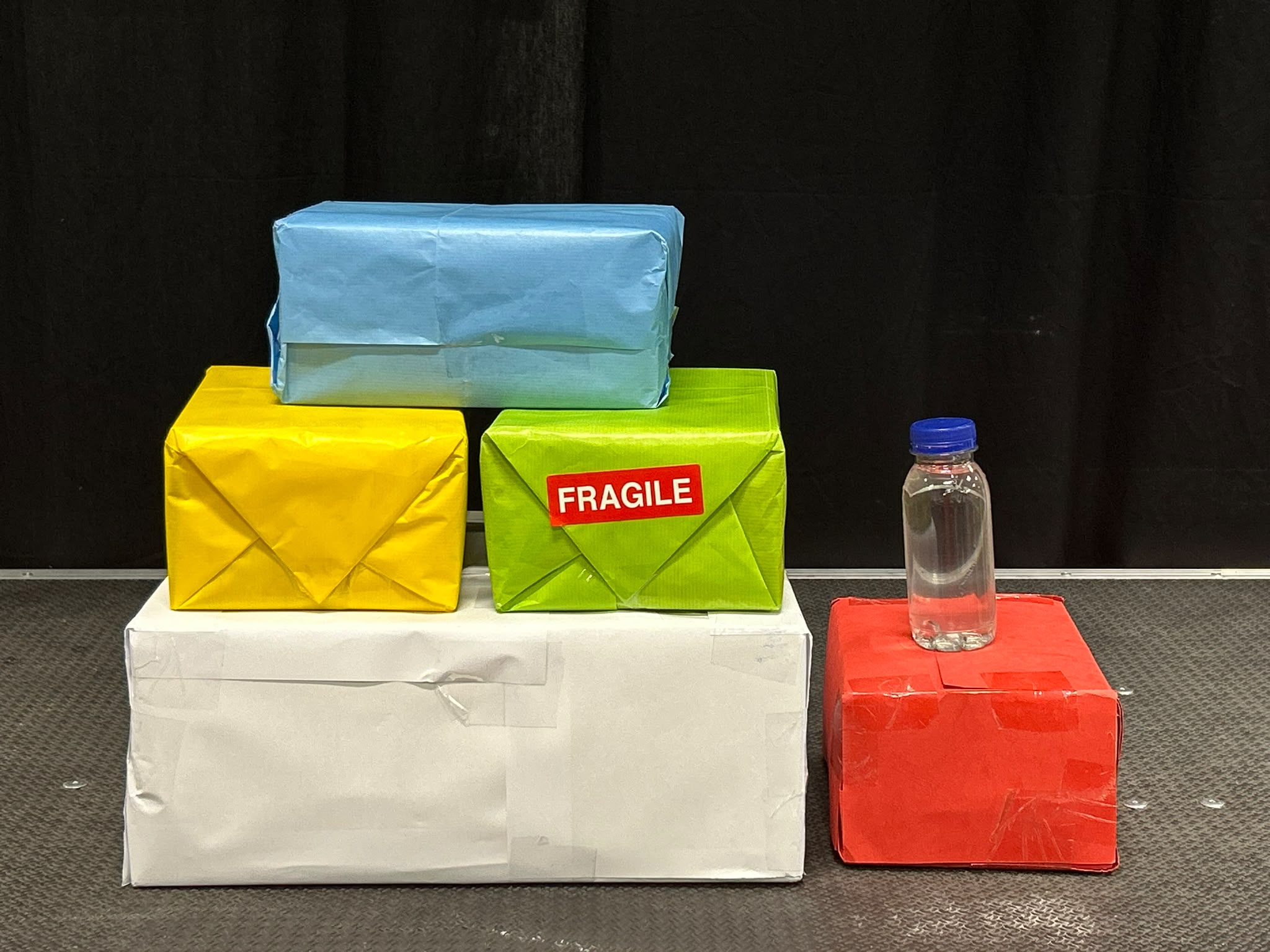}
        \caption{Long Horizon}
    \end{subfigure}

    \vspace{0.15cm}

    \begin{subfigure}[t]{0.32\linewidth}
        \centering
        \includegraphics[width=\linewidth]{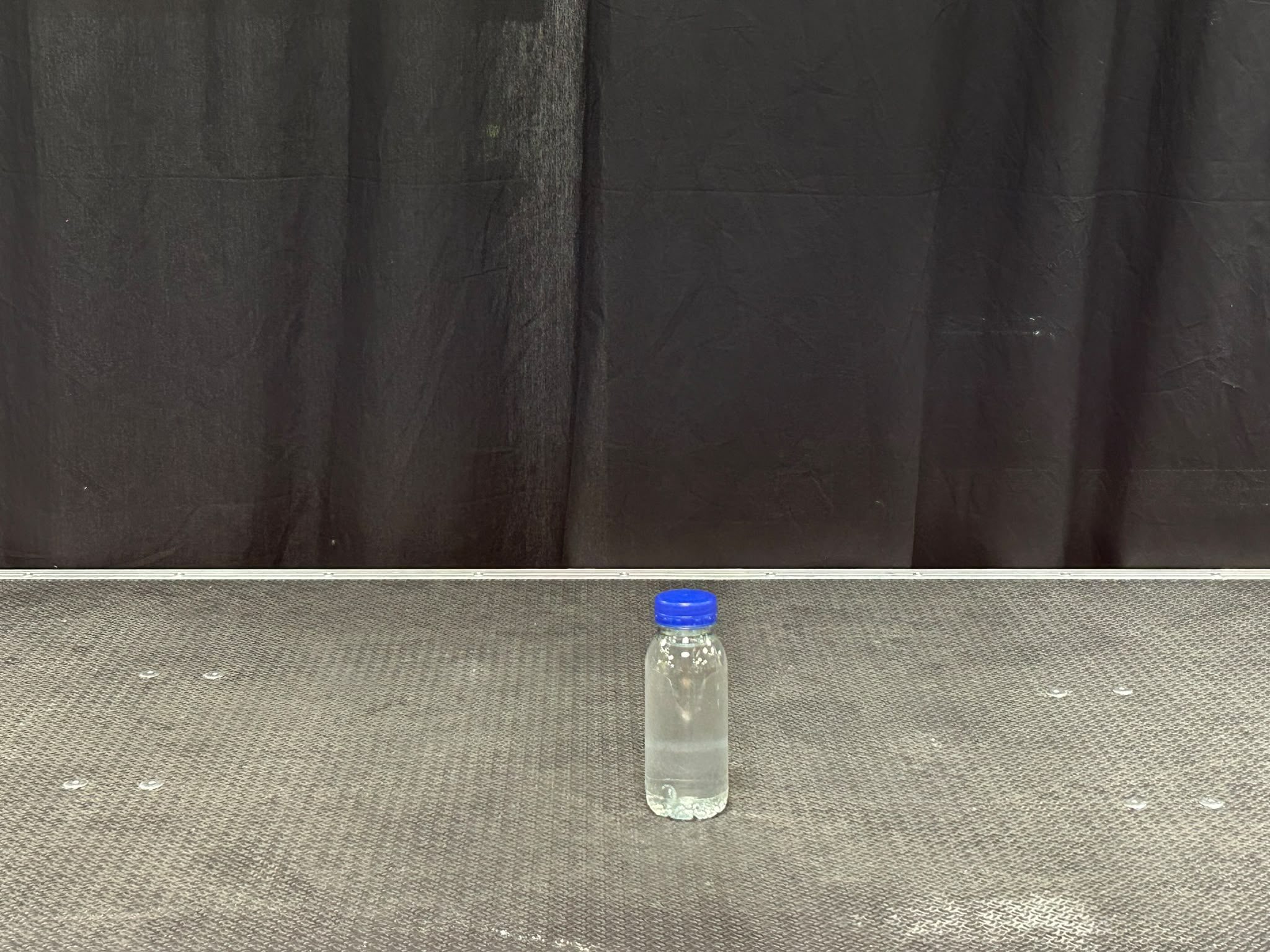}
        \caption{Bottle Failure}
    \end{subfigure}
    \hfill
    \begin{subfigure}[t]{0.32\linewidth}
        \centering
        \includegraphics[width=\linewidth]{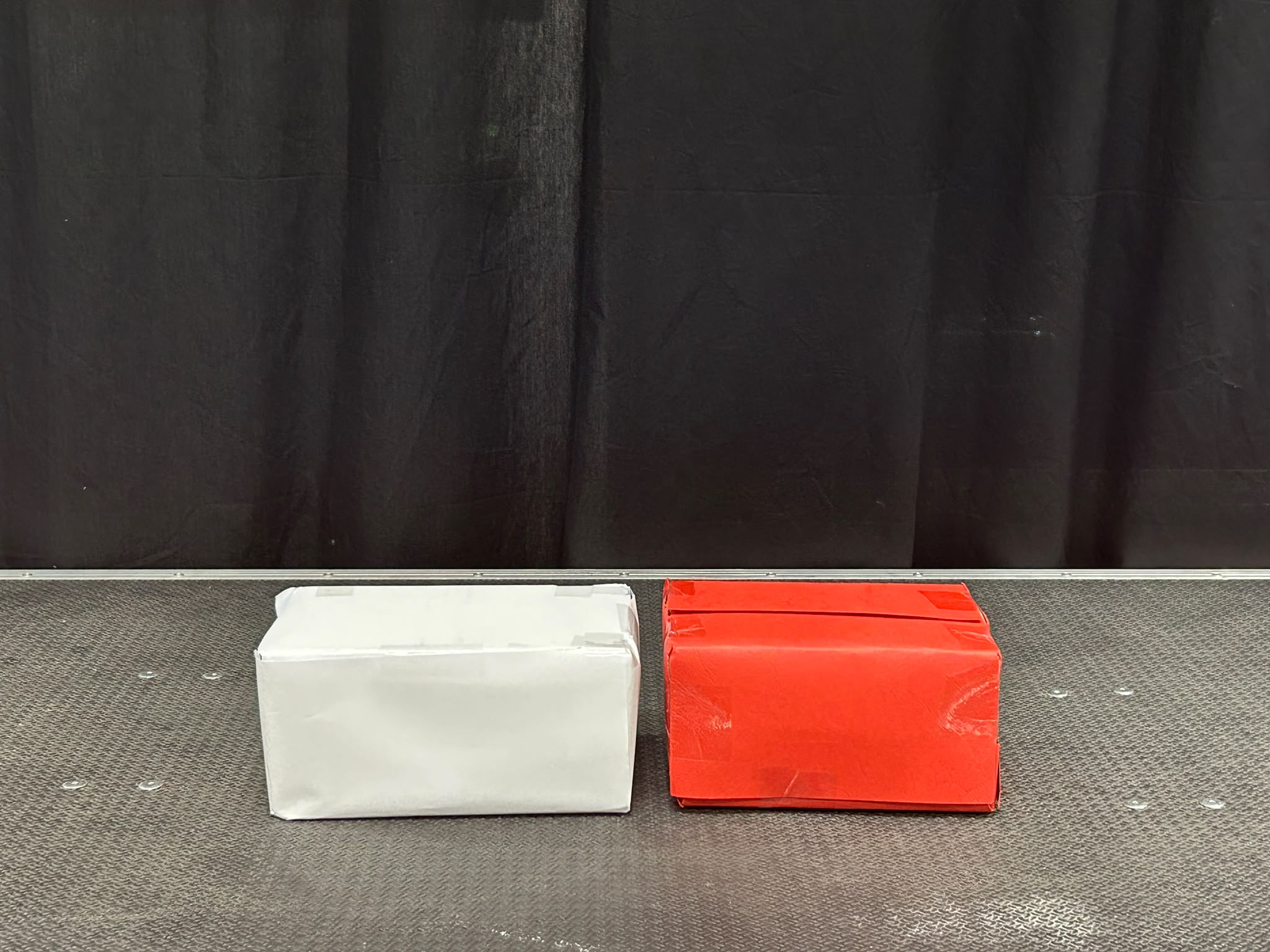}
        \caption{Push Failure}
    \end{subfigure}
    \hfill
    \begin{subfigure}[t]{0.32\linewidth}
        \centering
        \includegraphics[width=\linewidth]{images/stacked_boxes.jpeg}
        \caption{Stacked Boxes w/ Failure}
    \end{subfigure}
    
    \caption{Initial configurations of the nine tabletop manipulation scenarios used in the experimental evaluation.}
    \label{fig:scenarios}
\end{figure}

\begin{figure*}[t]
    \centering
    \includegraphics[width=1.0\linewidth]{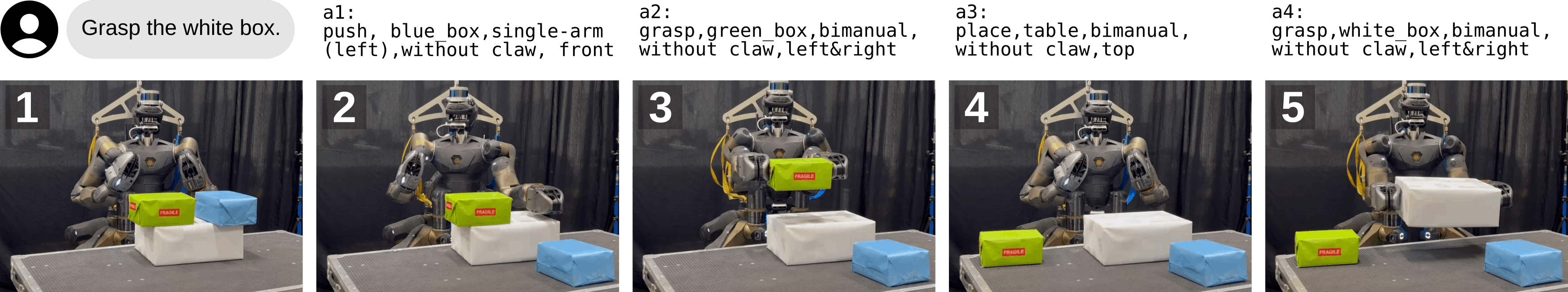}
    \caption{Example execution of the \emph{Fragile Box} scenario, where the task
is to grasp the white box. The robot identifies the blue and green boxes
as obstacles placed on top of the target. Since they are laterally too close to
be grasped directly, the robot first pushes the non-fragile blue box from its
accessible \emph{front} side, the most suitable direction for clearing the
target. It then grasps and places the fragile green box on the table. Once both
obstacles have been removed, the robot grasps the white box.}
    \label{fig:timeline}
\end{figure*}

\subsection{Scheduler}

The manipulation plan $\Pi_t$ is initially generated as a sequential list of actions. Since this representation does not explicitly exploit the robot's capability to execute compatible actions in parallel, a dedicated scheduling module reorganises the plan into a sequence of execution stages while preserving the original action semantics. Actions are grouped according to their compatibility, considering object interactions, spatial constraints, and the allocation of the robot end-effectors, ensuring that concurrent execution does not introduce conflicts. The resulting stage sequence is
\begin{equation}
S_t = (s_1,\dots,s_m),
\end{equation}
where each stage $s_i$ is a set of mutually compatible actions to be executed concurrently, e.g., $s_i=\{a_p,a_q\}$ with $a_p,a_q\in\Pi_t$.
For each stage, the scheduler additionally defines a \textit{local goal} $g_i$,
together with symbolic \textit{pre-conditions} $C_i^{pre}$ and \textit{post-conditions} $C_i^{post}$. The local goal summarises the intended outcome of the concurrent actions, while the symbolic conditions describe the required scene state before execution and the expected state afterwards. Rather than representing these conditions as complete scene descriptions, they are expressed as collections of atomic predicates, each describing a single independently verifiable property of the scene. By associating each group of compatible actions with a local goal and explicit symbolic conditions, the stage becomes the fundamental unit of supervision, providing the information required to assess the intended state transition. This representation naturally supports local recovery whenever possible, while global replanning is reserved for situations in which the remaining schedule can no longer be reconciled with the observed scene.

\subsection{Hierarchical Validation \& Recovery}
\label{subsec:validation_v2}
Each stage $s_i$ is supervised by a VLM-based validation module that compares
the scene evolution expected by the schedule with the one actually observed.
Before execution, the pre-conditions $C_i^{pre}$ are evaluated on the evidence
$Z_i^{pre}=\{I_i^{pre}\}$, given by the current observation. After execution,
the post-conditions $C_i^{post}$ are evaluated on
$Z_i^{post}=\{I_i^{pre}, I_i^{post}\}$ together with the executed actions, so
that the validator can verify that the expected changes were actually produced
by the stage. Unlike binary validation, the VLM assigns each condition
$c \in C_i^{\phi}$, with $\phi\in\{pre,post\}$, one of three outcomes
\begin{equation}
    v_{\phi}(Z_i^{\phi},c) \in
    \{\emph{satisfied}, \emph{uncertain}, \emph{violated}\}.
    \label{eq:validation_v2}
\end{equation}
These outcomes are aggregated into a stage-level outcome
$\gamma_i^{\phi}=\Gamma(\{v_{\phi}(Z_i^{\phi},c) \mid c\in C_i^{\phi}\})$,
where $\Gamma$ returns \emph{violated} if any condition is violated,
\emph{uncertain} if the VLM is unsure about at least one condition given
$Z_i^{\phi}$, and \emph{satisfied} otherwise. A \emph{satisfied} outcome allows
execution to proceed, whereas an \emph{uncertain} one is first addressed
through additional perception, by reconstructing the scene graph and
re-evaluating the conditions with it as additional evidence. Only when a failure is confirmed, or the ambiguity persists after
a limited number of attempts, is the stage forwarded to the last level of the
hierarchy, namely failure interpretation and recovery.

When a failure is detected, a structured report is built. 
If the 
pre-conditions are violated, the stage has not been executed and cannot be
repaired locally, so replanning is selected directly. If the post-conditions
are violated, the report collects the unsatisfied conditions with their
justifications, the executed actions, the
scene graphs reconstructed from $I_i^{pre}$ and $I_i^{post}$, and the history
$L_i$ of previous attempts on the stage. Comparing the two scene
graphs reveals whether execution made progress towards the local goal $g_i$
and whether the pre-conditions of $s_i$ still hold in the updated scene, while
$L_i$ exposes recurring failures and the strategies already applied.
Together, these elements indicate whether execution made partial progress,
whether the failure can be attributed to an action attribute, or whether the
stage is no longer applicable.

Recovery then follows a local-to-global policy, which attempts to repair the
current stage before revising the whole plan. \emph{Repeat} is selected when execution made progress
towards the local goal without completing it: the scene is perceived again and
the failed stage is re-executed unchanged.
\emph{Modify} is selected when the action is not completed and the failure can be attributed to an action parameter. In this case, the scene is perceived again and an \emph{Action Revision} VLM compares the observations of the previous attempts. Based on the failed conditions and the parameter values already tried in $L_i$, the VLM revises $\textit{mode}_k$, $\textit{contact}_k$, or $\textit{side}_k$ to address the observed failure. The stage is then re-executed with the revised actions.
\emph{Replan} is selected when the stage is no longer applicable in the updated
scene, when the evidence remains insufficient, or when neither local strategy
is supported. In this case, the scene is perceived again, and manipulation planning and
scheduling are re-invoked to regenerate the remaining plan. To avoid repeating
the same failure, the planner additionally receives a summary of the failed
stage, its violated conditions, and the strategies already tried in $L_i$. As summarised in
Table~\ref{tab:recovery}, the three strategies thus re-invoke progressively
larger portions of the pipeline. Moreover, \emph{abort}
terminates execution when no strategy remains within the allowed number of
attempts. Each attempt is appended to $L_i$, so that ineffective
strategies are not retried, and the task is considered complete when the
post-conditions of every stage have been validated as \emph{satisfied}.

\begin{table*}[t]
    \centering
    \caption{Scenario-level and module-level performance over 20 executions
    per scenario. Scenario performance: SR is the percentage of executions
    completed successfully, and Queries is the average number of VLM queries per
    successful execution (--: no successful execution). Module performance
    (Ours) is computed over all the calls of each module: SR is the percentage
    of calls in which the module produces a correct output, and Latency is the
    average time of a single call, in seconds. For Hierarchical Validation,
    Latency is the average between pre-execution and post-execution validation.
    Values in bold denote the best results.}
    \label{tab:success_rates}

    \begingroup
    \setlength{\tabcolsep}{4pt}
    \renewcommand{\arraystretch}{1.15}
    \footnotesize

    \resizebox{\textwidth}{!}{%
    \begin{tabular}{l*{14}{c}}
        \toprule

        & \multicolumn{6}{c}{\textbf{Scenario performance}}
        & \multicolumn{8}{c}{\textbf{Module performance (Ours)}} \\
        \cmidrule(lr){2-7}
        \cmidrule(lr){8-15}

        & \multicolumn{2}{c}{
            \shortstack{\textbf{Non-execution-aware \&}\\
                        \textbf{Binary validation}}}
        & \multicolumn{2}{c}{
            \shortstack{\textbf{Execution-aware \&}\\
                        \textbf{Binary validation}}}
        & \multicolumn{2}{c}{
            \shortstack{\textbf{Execution-aware \&}\\
                        \textbf{Hierarchical validation (Ours)}}}
        & \multicolumn{2}{c}{
            \shortstack{\textbf{Scene}\\
                        \textbf{Perception}}}
        & \multicolumn{2}{c}{
            \shortstack{\textbf{Manipulation}\\
                        \textbf{Planning}}}
        & \multicolumn{2}{c}{
            \shortstack{\textbf{Execution}\\
                        \textbf{Scheduling}}}
        & \multicolumn{2}{c}{
            \shortstack{\textbf{Hierarchical}\\
                        \textbf{Validation}}} \\

        \cmidrule(lr){2-3}
        \cmidrule(lr){4-5}
        \cmidrule(lr){6-7}
        \cmidrule(lr){8-9}
        \cmidrule(lr){10-11}
        \cmidrule(lr){12-13}
        \cmidrule(lr){14-15}

        \textbf{Scenario}
        & SR & Queries
        & SR & Queries
        & SR & Queries
        & SR & Latency
        & SR & Latency
        & SR & Latency
        & SR & Latency \\
        \midrule

        \multicolumn{15}{l}{\textit{Execution-Aware Planning}} \\

        \quad Stacked Boxes
        & 100\% & 8.2
        & 100\% & 8.2
        & \textbf{100\%} & \textbf{8.2}
        & 100\% & 5.10
        & 100\% & 9.69
        & 100\% & 8.42
        & 100\% & 5.85 \\

        \quad Boxes Aside
        & 0\% & --
        & 75\% & 12.3
        & \textbf{85\%} & \textbf{11.8}
        & 96\% & 7.33
        & 92\% & 12.82
        & 100\% & 8.32
        & 98\% & 4.45 \\

        \quad Bottles on Box
        & 70\% & 9.6
        & 90\% & 9.4
        & \textbf{95\%} & \textbf{9.4}
        & 98\% & 7.78
        & 100\% & 10.93
        & 85\% & 10.01
        & 97\% & 4.91 \\

        \quad Fragile Box
        & 0\% & --
        & 70\% & 11.7
        & \textbf{70\%} & \textbf{11.5}
        & 90\% & 6.72
        & 84\% & 22.30
        & 100\% & 11.52
        & 97\% & 4.76 \\

        \quad Long Horizon
        & 0\% & --
        & 35\% & 24.3
        & \textbf{50\%} & \textbf{21.1}
        & 78\% & 8.06
        & 67\% & 31.78
        & 91\% & 14.29
        & 93\% & 4.57 \\

        \addlinespace
        \multicolumn{15}{l}{\textit{Dynamic Scene Adaptation}} \\

        \quad Scene Update
        & 85\% & 14.6
        & 85\% & 14.6
        & \textbf{85\%} & \textbf{14.4}
        & 97\% & 7.64
        & 96\% & 8.23
        & 100\% & 12.54
        & 97\% & 5.60 \\

        \addlinespace
        \multicolumn{15}{l}{\textit{Execution Recovery}} \\

        \quad Bottle Failure
        & 90\% & 10
        & 90\% & 10
        & \textbf{90\%} & \textbf{8.1}
        & 100\% & 4.73
        & 100\% & 5.79
        & 100\% & 3.55
        & 92\% & 4.34 \\

        \quad Push Failure
        & 15\% & 10
        & 85\% & 10
        & \textbf{95\%} & \textbf{9.1}
        & 98\% & 6.38
        & 100\% & 9.29
        & 100\% & 6.64
        & 89\% & 6.52 \\

        \quad Stacked Boxes w/ Failure
        & 90\% & 12
        & 90\% & 12
        & \textbf{90\%} & \textbf{11.8}
        & 100\% & 4.52
        & 95\% & 9.46
        & 100\% & 8.28
        & 91\% & 5.19 \\

        \bottomrule
    \end{tabular}%
    }

    \endgroup
\end{table*}

\section{Experiments}

\subsection{Experimental Setup}

Experiments are conducted on the real CENTAURO hybrid wheeled-legged humanoid platform~\cite{kashiri2019centauro}, whose two six-degree-of-freedom arms are equipped with Dagana end-effectors~\cite{del2024high}. Manipulation actions follow joint-space references computed by a QP-based inverse kinematics module~\cite{laurenzi2019cartesi}, while the base remains stationary throughout all experiments. Scene observations are acquired from the RGB stream of a head-mounted Intel RealSense D435 camera directed towards the manipulation workspace. Unless otherwise stated, all VLM-based modules use Gemini Robotics-ER 1.6.
The framework is evaluated on nine tabletop manipulation scenarios, shown in Fig.~\ref{fig:scenarios}, covering execution-aware planning, adaptation to scene changes, and execution recovery. Each scenario is executed 20 times from the same initial configuration. Planning scenarios address increasing task complexity, from accessibility-aware retrieval (\textit{Stacked Boxes} and \textit{Boxes Aside}) to coordinated bimanual manipulation (\textit{Bottles on Box}), semantic reasoning (\textit{Fragile Box}), and long-horizon planning (\textit{Long Horizon}). \textit{Scene Update} introduces unexpected environmental changes that invalidate the original plan, while \textit{Bottle Failure}, \textit{Push Failure}, and \textit{Stacked Boxes w/ Failure} evaluate hierarchical validation under unsuccessful grasp attempts, manipulation deviations, and absence of observable execution progress, respectively.

\subsection{Results and Analysis}
\label{sec:overall_evaluation}

In this section, we first compare the complete framework with two baselines that share the same
VLM backbone, task instructions, manipulation primitives, and low-level
execution modules. The first baseline (\textit{Non-execution-aware \& Binary
validation}) provides the planner only with symbolic object relations, without
side-level accessibility and reachability information, and relies on a binary
success/failure validator. The second (\textit{Execution-aware \& Binary
validation}) adopts the proposed execution-aware scene representation but
retains the binary validator. The proposed framework (\textit{Execution-aware \&
Hierarchical validation}) additionally replaces the binary validator with
hierarchical validation. Furthermore, we analyse the role of the
individual modules, the benefit of parallel execution, and the choice of the VLM
backbone.

\subsubsection{Overall Performance and Baselines}

\begin{figure}[t]
\centering
\definecolor{recRepeat}{HTML}{4C84B8}
\definecolor{recModify}{HTML}{F5A81C}
\definecolor{recReplan}{HTML}{8A97A6}
\begin{tikzpicture}
\pgfplotsset{recaxis/.style={
    scale only axis, width=0.66\columnwidth, height=3.9cm,
    xmin=0, xmax=3.8, ymin=-0.6, ymax=5.6, y dir=reverse}}
\begin{axis}[recaxis,
    xbar stacked,
    bar width=11pt,
    xtick={0,1,2,3},
    xlabel={VLM queries per recovery decision},
    axis x line*=bottom,
    axis y line*=left,
    y axis line style={draw=none},
    ytick={0,1,2,3,4,5},
    yticklabels={{Boxes Aside},{Long Horizon},{Scene Update},{Bottle Failure},{Push Failure},{Stacked Boxes\\w/ Failure}},
    yticklabel style={font=\footnotesize, align=right},
    major y tick style={draw=none},
    tick label style={font=\footnotesize},
    label style={font=\footnotesize},
    xmajorgrids=true,
    grid style={draw=black!12, line width=0.3pt},
    axis line style={draw=black!60},
    legend style={at={(0.5,1.04)}, anchor=south, legend columns=-1,
                  draw=none, font=\footnotesize,
                  /tikz/every even column/.append style={column sep=6pt}},
    area legend,
]
\addplot[fill=recRepeat, draw=white, line width=0.6pt] coordinates {(0.500,0) (0.333,1) (0.118,2) (0.944,3) (0.053,4) (0.056,5)};
\addplot[fill=recModify, draw=white, line width=0.6pt] coordinates {(1.000,0) (0.667,1) (0.000,2) (0.111,3) (1.684,4) (0.222,5)};
\addplot[fill=recReplan, draw=white, line width=0.6pt] coordinates {(0.000,0) (1.000,1) (2.647,2) (0.000,3) (0.316,4) (2.500,5)};
\legend{Repeat, Modify, Replan}
\end{axis}
\begin{axis}[recaxis, hide axis]
\draw[dashed, line width=0.8pt, black!70] (axis cs:3,-0.6) -- (axis cs:3,5.6);
\node[font=\scriptsize, anchor=west, inner sep=2pt, text=black!70] at (axis cs:3.03,2.5) {Binary};
\node[font=\scriptsize, anchor=west, inner sep=2pt] at (axis cs:1.500,0) {1.5};
\node[font=\scriptsize, anchor=west, inner sep=2pt] at (axis cs:2.000,1) {2.0};
\node[font=\scriptsize, anchor=east, inner sep=2pt] at (axis cs:2.765,2) {2.8};
\node[font=\scriptsize, anchor=west, inner sep=2pt] at (axis cs:1.056,3) {1.1};
\node[font=\scriptsize, anchor=west, inner sep=2pt] at (axis cs:2.053,4) {2.1};
\node[font=\scriptsize, anchor=east, inner sep=2pt] at (axis cs:2.778,5) {2.8};
\end{axis}
\end{tikzpicture}
\caption{Average number of VLM queries spent per recovery decision in six representative
scenarios, considering the successful executions. Each bar is the average over
the decisions of the hierarchical validator, and its coloured segments show the contribution of repeat (1 query),
modify (2), and replan (3), weighted by how often each is selected. The dashed
line is the binary validator, which always replans (3 queries per failure). Only
the queries of the recovery decisions are counted, not the validation calls of
the repeated stages.}
\label{fig:recovery_cost}
\end{figure}
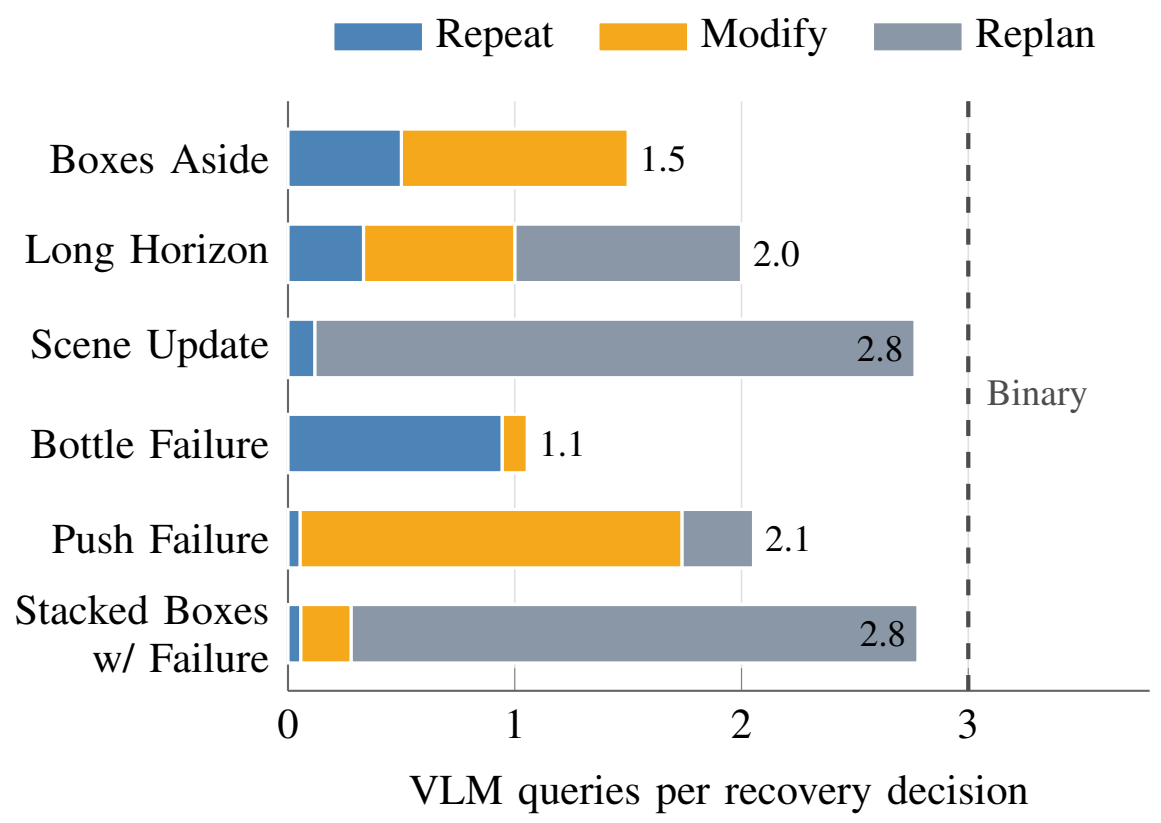

Table~\ref{tab:success_rates} reports the performance of the three
configurations on the nine scenarios.
The benefit of execution awareness depends on how objects obstruct the target.
When an obstacle rests on top of the target, as in \textit{Stacked Boxes}, this
dependency is already captured by the symbolic relations, and the
non-execution-aware planner also solves the scenario. When objects are placed
side by side, as in \textit{Boxes Aside}, \textit{Fragile Box}, and
\textit{Long Horizon}, however, the symbolic representation cannot tell whether
the free space between them is sufficient for the end-effector to reach the
target. The non-execution-aware planner therefore does not recognise a
laterally placed object as an obstruction and attempts to grasp the target
directly, failing in all executions (0\% SR). By encoding side-level
accessibility and reachability, the execution-aware planner instead identifies
and relocates the blocking objects, reaching 85\%, 70\%, and 50\% SR.
Fig.~\ref{fig:timeline} shows an example execution of \textit{Fragile Box}.
Similar gains appear in \textit{Bottles on Box} (from 70\% to 95\%) and
\textit{Push Failure} (from 15\% to 95\%), suggesting that these scenarios
rely on the same interaction-relevant information.
The main contribution of hierarchical validation lies in efficiency. A binary
validator only reports whether a stage succeeded and, without information on the
cause of a failure, must always replan, spending 3 queries per failure. The
hierarchical validator instead interprets the failure and selects the least
disruptive recovery. Fig.~\ref{fig:recovery_cost} reports the average cost of a recovery
decision in six scenarios. In \textit{Bottle Failure}, the grasp fails once per
execution: the binary validator replans from scratch, spending 3 queries,
whereas the hierarchical validator mostly just repeats the action, so a decision
costs 1.1 queries on average. These 1.9 saved queries correspond to the drop
from 10 to 8.1 queries per execution in Table~\ref{tab:success_rates}. In
\textit{Push Failure}, repeating is not enough but a modification suffices, which
costs 2.1 queries and lowers the queries from 10 to 9.1. When the failure
invalidates the plan, as in \textit{Stacked Boxes w/ Failure} and
\textit{Scene Update}, replanning is almost always unavoidable, and both validators require a
similar number of queries.

\subsubsection{Module-level Analysis}
The module-level columns of Table~\ref{tab:success_rates} break down the
performance of the proposed framework by module. Execution scheduling achieves
the highest SR, equal to 100\% in seven of the nine scenarios, since its task is
limited to grouping compatible actions and its input is purely textual.
Manipulation planning also receives only textual input, but its SR depends
strongly on the task: generating a plan with many steps and heterogeneous
interactions, as in \textit{Long Horizon} (67\%), is considerably harder than
solving a scenario such as \textit{Stacked Boxes} (100\%). Scene perception and validation, instead, take both text and
images as input, and their failures stem less from the task than from scene
ambiguities, perspective effects, and suboptimal viewpoints caused by the
head-mounted camera.
The same pattern emerges from the latencies. Scene perception and validation
have fairly uniform latencies, whereas planning and, to a lesser extent,
scheduling scale with the complexity of the task. In particular, the planning
latency grows from 5.79--12.82\,s in the simpler scenarios to 22.30\,s in
\textit{Fragile Box} and 31.78\,s in \textit{Long Horizon}.

\subsubsection{Benefit of Parallel Execution}

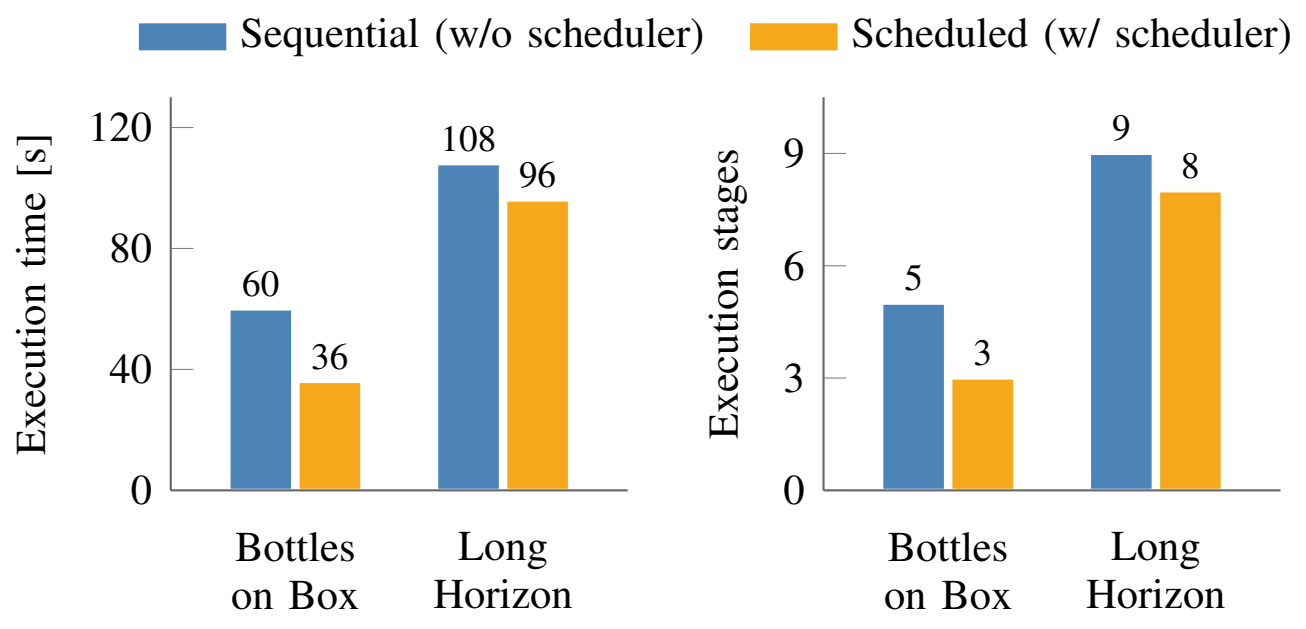
\begin{figure}[t]
\centering
\definecolor{schSeq}{HTML}{4C84B8}
\definecolor{schPar}{HTML}{F5A81C}
\begin{tikzpicture}
\pgfplotsset{schaxis/.style={
    scale only axis, width=0.35\columnwidth, height=2.6cm,
    ybar, bar width=12pt, ymin=0,
    xmin=-0.6, xmax=1.6, xtick={0,1},
    xticklabels={{Bottles\\on Box},{Long\\Horizon}},
    xticklabel style={align=center, font=\footnotesize},
    tick label style={font=\footnotesize}, label style={font=\footnotesize},
    major x tick style={draw=none},
    axis x line*=bottom, axis y line*=left, axis on top,
    axis line style={draw=black!60},
    nodes near coords,
    every node near coord/.append style={font=\scriptsize, inner sep=2pt},
    anchor=south west}}
\begin{axis}[schaxis, at={(0.11\columnwidth,0)},
    ymax=130, ytick={0,40,80,120}, ylabel={Execution time [s]},
    legend style={at={(1.2,1.05)}, anchor=south, legend columns=-1,
                  draw=none, font=\footnotesize,
                  /tikz/every even column/.append style={column sep=6pt}},
    area legend,
]
\addplot[fill=schSeq, draw=white, line width=0.6pt, bar shift=-6.5pt] coordinates {(0,60) (1,108)};
\addplot[fill=schPar, draw=white, line width=0.6pt, bar shift=6.5pt]  coordinates {(0,36) (1,96)};
\legend{Sequential (w/o scheduler), Scheduled (w/ scheduler)}
\end{axis}
\begin{axis}[schaxis, at={(0.61\columnwidth,0)},
    ymax=10.5, ytick={0,3,6,9}, ylabel={Execution stages},
]
\addplot[fill=schSeq, draw=white, line width=0.6pt, bar shift=-6.5pt] coordinates {(0,5) (1,9)};
\addplot[fill=schPar, draw=white, line width=0.6pt, bar shift=6.5pt]  coordinates {(0,3) (1,8)};
\end{axis}
\end{tikzpicture}
\caption{Sequential execution versus the proposed bimanual scheduler on the
\textit{Bottles on Box} and \textit{Long Horizon} scenarios, in terms of
execution time (left), which accounts only for the robot motion and not for the
model queries, and number of execution stages (right).}
\label{fig:scheduler_analysis}
\end{figure}

We further compare sequential execution with the proposed bimanual scheduler,
which groups compatible actions into parallel stages, on the
\textit{Bottles on Box} and \textit{Long Horizon} scenarios
(Fig.~\ref{fig:scheduler_analysis}). The reported values are not averages over
the 20 runs, but highlight the efficiency that scheduling can achieve in these
scenarios, and the execution time accounts only for the robot motion, not for the
model queries. Parallel scheduling reduces the execution time by 40\% in
\textit{Bottles on Box} and by 11\% in \textit{Long Horizon}, and the number of
execution stages from 5 to 3 and from 9 to 8, respectively, while preserving
task completion.

\begin{table}[t]
\centering
\caption{Comparison of different VLM backbones within the proposed framework,
averaged over the \textit{Boxes Aside}, \textit{Scene Update}, and
\textit{Long Horizon} scenarios. SR is the percentage of executions completed
from start to end, and Latency is the sum of the average latencies of scene
perception, manipulation planning, execution scheduling, and validation.}
\label{tab:backbone_comparison}

\begingroup
\setlength{\tabcolsep}{4pt}
\renewcommand{\arraystretch}{1.1}
\footnotesize

\begin{tabular}{lcc}
\toprule
\textbf{Backbone}
&
\textbf{SR}
&
\textbf{Latency}
\\
\midrule

Gemini Robotics-ER 1.6
&
73.3\%
&
41.9\,s
\\

GPT-5.2
&
5.0\%
&
34.6\,s
\\

GPT-o3
&
80.0\%
&
91.4\,s
\\

\bottomrule
\end{tabular}

\endgroup
\end{table}

\subsubsection{Backbone Sensitivity}

Finally, to assess the sensitivity of the framework to the underlying VLM, we
replace Gemini Robotics-ER 1.6 with GPT-o3 and GPT-5.2 as backbone under
identical conditions. Table~\ref{tab:backbone_comparison} reports the results on
three representative scenarios, each executed 20 times. Gemini and o3 achieve comparable SR (73.3\% and
80.0\%), with o3 requiring a considerably higher latency (91.4\,s versus
41.9\,s). GPT-5.2 reaches only 5.0\% despite its lower latency
(34.6\,s). Its failures arise mainly in the manipulation planning stage,
where the model struggles to produce feasible long-horizon plans, either because
the logic of the action sequence is incorrect or because the parameters of
individual actions are inconsistent.

\section{Conclusion and Future Work}
This work presented a zero-shot, language-conditioned framework for execution-aware bimanual manipulation in cluttered environments. By explicitly encoding blocking relations, side-level accessibility, and kinematic reachability, the proposed framework reasons jointly about task objectives and their physical realisation. Furthermore, the proposed hierarchical validation strategy explicitly relates execution feedback to planning decisions, enabling targeted recovery and replanning when the observed scene evolution deviates from the expected one. Experimental results on the real CENTAURO platform demonstrate that combining structured scene understanding, execution-aware planning, and closed-loop validation improves the robustness of language-conditioned manipulation in cluttered environments. 
Promising directions for future work include relaxing these assumptions.
Learned manipulation policies could replace inverse kinematics to enable more
adaptive and contact-rich behaviours, while more flexible skill generation and
composition could extend the set of available actions. Moreover, reachability
is currently evaluated from a fixed base pose, so kinematically unreachable
sides are simply discarded; jointly planning manipulation and base motion would
allow the robot to reposition itself and turn these constraints into additional
planning decisions.

\bibliographystyle{IEEEtran}
\bibliography{bibliography}

\end{document}